\documentclass[letterpaper, 11pt, oneside]{Thesis}  

\usepackage{verbatim}  
\usepackage{amsmath}
\usepackage{graphicx}
\usepackage{caption}
\usepackage{calc}
\usepackage[algo2e]{algorithm2e} 
\usepackage{algorithm}
\usepackage{pifont}
\usepackage{vector}  
\usepackage[letterpaper, margin=1.9in]{geometry}
\hypersetup{urlcolor=blue, colorlinks=true}  
\newcommand\norm[1]{\left\lVert#1\right\rVert}
\usepackage{titlesec}

\setmarginsrb{30mm}{30mm}{30mm}{30mm}{10pt}{7mm}{10pt}{5mm}
\titleformat{\chapter}[display]
  {\normalfont\bfseries}{}{0pt}{\Huge}

\begin{document}
\begin{titlepage}
\pagenumbering{gobble}
\begin{center}
%
  \vskip.1in
  \textbf{\huge Visual Based Navigation of Mobile Robots}  \vskip.7in
  \vskip.7in
	\begin{center}\textit{\small{Report submitted in partial fulfillment to the requirements}} \\
	\textit{\small{for the degree of}}
	\vskip.5in
	\large{\normalsize BACHELOR OF TECHNOLOGY}
	\vskip.3in
	\textit{by}
	\vskip.3in
	
	\textbf{\normalsize SHAILJA}\\
	\end{center}
\end{center}
%
\vskip.9in
\vskip.9in

\begin{center}
\includegraphics[scale=0.18]{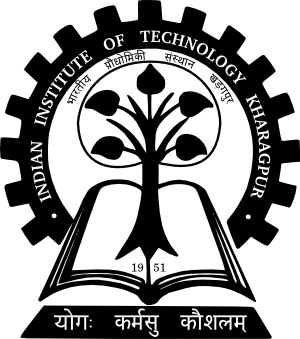}
\vskip.1in
\textbf{Department of Electrical Engineering}
\\
Indian Institute of Technology, Kharagpur
\\
West Bengal, INDIA 721302\\
November 2015

\end{center}
\end{titlepage}


\setlength{\parindent}{85pt}
\renewcommand{\headrulewidth}{0pt}
\begin{center} \huge{\textbf{Certificate}} \end{center}
\vskip.4in
{\setlength{\parindent}{0cm}This is to certify that the report entitled \textbf{Visual Based Navigation of Mobile Robots}, submitted by \textbf{Shailja}, an undergraduate student, in the \textit{Department of Electrical Engineering, Indian Institute of Technology,
Kharagpur, India,} for the award of the degree of Bachelor of Technology, is a record of
the work carried out by her under our supervision and guidance. Neither this report nor any part of it has been submitted for any degree or academic award elsewhere.}
\vskip.9in
\vskip.5in
\parindent=0em
\begin{minipage}[p]{10cm}
\begin{flushleft}
\textbf{Dr. Jayanta Mukhopadhyay}
\\Professor,\\
Department of Computer Science and Engineering,\\ Indian Institute of Technology,
Kharagpur,\\ INDIA 721302
\end{flushleft}
\end{minipage}

\vskip.9in
\vskip.9in
\vskip.9in
\vskip.9in

\pagestyle{fancy}  

\tableofcontents  

\mainmatter
\renewcommand{\headrulewidth}{1pt}
\pagestyle{fancy}  


\chapter{Introduction}

\section{Project Overview}
As humans, if a mobile robot has to get familiar with a completely unknown environment, it should be able to build the map of the environment by exploring it. Like us, Robots can map only a part of the environment which is visible from current position. It should be able to navigate autonomously and planning a safe obstacle free path to reach there.
This finds applications in personal assistant robot which can carry out tasks thereby saving us time and improving the quality of our lives.  Moreover, some tasks can be completed more effectively by a machine rather than a person.  Life would be so much easier and richer with an intelligent personal robot in the house! In  contrast  to  industrial  robots,  which  are  generally  programmed  to  carry  out  a  few fixed tasks, personal robots need to adapt to wide variety of human environments and carry  out  a  diverse  array  of  tasks,  learning  as  they  go.   In  light  of  this,  we  see  that the basic skills of understanding the environment such as mapping and perception are
critical for such robots. 

It  is  quite  a  challenging  task  to  develop  such  an autonomous system. \cite{pant}
With improvements in technology and the vast amount of research being done in the
field  of  robotics,  a  few  mobile  robots  have  already  been  built  that  utilize  laser  range finders  as  their  primary  sensor  to  operate. However,  one  of  the  major  hurdles standing in the way of widespread household adoption is the cost.  All these robots are very expensive and in general, most people will not be able to afford them.

In this project we are applying algorithms which can be helpful in obstacle free navigation by detecting real world co-ordinates and suitable window of advancement.
\section{Challenges and Constraints}

Autonomous robot navigation itself is a challenging task. On top of this, making one with just monocular vision seems to be a daunting task! Sonar sensors have low range and resolution while laser range finders offer good resolution and in general produce good quality point clouds. They provide reasonably accurate distance measurements and obstacle detection algorithms have been developed using this range data. Moreover, almost all of the
existing Simultaneous Localization and Mapping (SLAM) \cite{slam} based algorithms rely on range data to function. Instead of expensive laser sensors, a stereo vision system can possibly be used to get the depth information by comparing the disparity between the left and right images. Two cameras are also much less expensive than a laser range finder. However, stereo vision systems have their own shortcomings. They often produce very sparse point clouds, require careful calibration, and moreover are computationally quite expensive (disparity map and optical flow calculation). The major shortcoming of monocular vision is the loss of this depth data since the points traced along a ray from the camera are mapped into one pixel. New methods need to be developed to either bypass this requirement, or to estimate the depth. One big advantage of vision is that it allows the robot to “see” and hence “understand”. With just a distance sensor it would be “blind”, in the sense that it would not be able to recognize most of the objects lying around. With the advancement in object detection research, this could become a real possibility using monocular vision.

\chapter{Obstacle Detection}
Collision-free navigation, and hence obstacle detection, is a very important task for
autonomous mobile robots. Typically, most robots rely on range data such as that from
ultrasonic sensors, laser range finders or stereo vision to
detect obstacles. These sensors, especially the laser range finder, produce good results.
However, they have some major drawbacks. The laser range finder is expensive, and
hence would not be suitable for consumer use or for light-weight robots. Ultrasonic
sensors are cheaper, but they generally suffer from low angular resolution. Stereo vision
based approaches are computationally expensive, generally produce a sparse point cloud
and require precise calibration. Moreover, range data based approaches are unable to
distinguish between different surfaces of the same height (e.g. between pavement and
rocky areas in a park) or small/flat objects lying on the ground.
So, with the advancement in vision algorithms in the recent years, a monocular vision
based approach presents a good alternative. The key difference in monocular vision
based obstacle detection and range data based obstacle detection is that in the former,
obstacles are distinguished from the ground by their appearance, whereas in the latter,
they are detected by the difference in their relative distance. We present a  multi-stage method for obstacle detection.
\section{Overview}
The obstacle detection algorithm takes a single image from the robot’s camera feed, and determines which areas are traversable and which parts are obstacles. We have made the following assumptions for speed and simplicity:
\begin{itemize}
\item The area immediately in front of the robot is free of obstacles.
\item All obstacles have their base on the ground, i.e. there are no “hanging” obstacles.
\item  Robot motion is constrained to the horizontal ground plane.

\end{itemize}
The first step consists of segmenting the input image into superpixels, which are local groupings of ”similar” pixels. Simple Linear Iterative Clustering is a fast algorithm used for finding the superpixels in an image. Superpixel segmentation allows the image to be processed as a set of regions, rather than a set of individual pixels, which speeds up the subsequent stages. The second step consists of histogram-based sampling. The superpixels that represent the immediate vicinity of the robot are sampled, and the color information is stored in a histogram. Next, we run a breadth first search over the superpixels, and classify regions as traversable or non-traversable in the process according to a membership criterion. Lastly, we estimate the floor junctions, where the obstacles meet the floor, from the camera image and use this to mask the final obstacle image. Fig. 2.1 shows a block diagram of the process.

\begin{figure}
\centering
\includegraphics{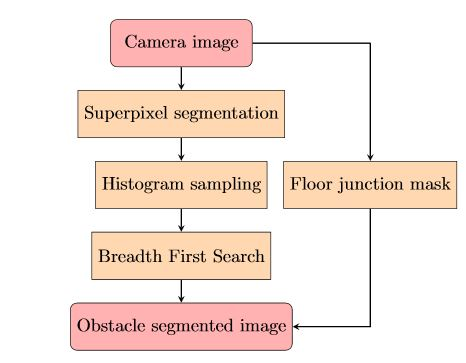}
\caption{Block diagram of the obstacle detection process}
\label{Fig. 2.1}
\end{figure}

\chapter{Superpixel Segmentation}

\section{Introduction}
Generally, images are represented as grid of pixels but pixel is not the natural representation of image. Pixel does not carry any semantic meaning of the corresponding image but analysing a local group of pixels will give more meaning. This local grouping of pixels is called a superpixel.
\section{Benefits of superpixel over pixel}
\begin{itemize}
\item Computational efficiency: While it may be computationally expensive to compute the actual superpixel groupings, superpixel allows us to reduce the complexity of the images themselves from hundreds of thousands of pixels to only a few hundred superpixels. Each of these superpixels will then contain some sort of perceptual, and ideally, semantic value.
\item Perceptual meaningfulness: Instead of only examining a single pixel in a pixel grid, which caries very little perceptual meaning, pixels that belong to a superpixel group share some sort of commonality, such as similar color or texture distribution.
\item Oversegmentation: Superpixel algorithms oversegment the image. This means that most of important boundaries in the image are found; however, at the expense of generating many insignificant boundaries. The end product of this oversegmentation is that that very little (or no) pixels are lost from the pixel grid to superpixel mapping.
\item Graphs over superpixel: Suppose if we want to represent an image as graph where each pixel represent a node and are related to other pixel through edges, this leads to a very complex representation with more number of pixels.
\end{itemize}
\section{Simple Linear Iterative Clustering (SLIC)}
We used SLIC algorithm for superpixel segmentation that clusters pixels in  the  combined  five-dimensional  color  and  image  plane  space  to  efficiently generate compact, nearly uniform superpixels. SLIC is simple to implement and easily applied in practice- the only parameter required is desired number of superpixels \cite{SLIC}.
\subsection{SLIC segmentation algorithm}
SLIC generates superpixels by clustering pixels based on their color similarity and proximity in the image. This can be done in five dimensional [rgbxy] space or [labxy] space, where [rgb] and [labxy] are the pixel vectors and xy is the pixel position. Superpixel size varies with the image size. It is not possible to simply use Euclidean distance in this 5D space without normalizing the spatial distances. A new distance measure that considers superpixel size is used which enforces color similarity as well as pixel proximity in this 5D space.
For an image with N pixels, if we desire K superpixels, then approximate size of each superpixel will be N/K pixels and for roughly equally sized superpixels there would be a superpixel center at every grid interval ${S=\sqrt{(N/K)}}$. \\
For RGB color space:
\begin{equation}
{d(p_{1} , p_{2})= \sqrt{ \big(r_{1}-r_{2}\big) ^{2}+\big(g_{1}-g_{2}\big) ^{2} +\big(b_{1}-b_{2}\big) ^{2} } + \frac{M}{S}\sqrt{ \big(x_{1}-x_{2}\big) ^{2}+\big(y_{1}-y_{2}\big) ^{2}} }
\label{dis}
\end{equation}
For CIELAB color space:\\
\begin{equation}
{d(p_{1} , p_{2})= \sqrt{ \big(l_{1}-l_{2}\big) ^{2}+\big(a_{1}-a_{2}\big) ^{2} +\big(b_{1}-b_{2}\big) ^{2} } + \frac{M}{S}\sqrt{ \big(x_{1}-x_{2}\big) ^{2}+\big(y_{1}-y_{2}\big) ^{2}}}
\end{equation}

{A variable M is introduced to control compactness of superpixel. The greater the value of M,, the more spatial proximity is emphasized and more compact the cluster. we have chosen M=10. K regularly spaced cluster centers and moving them to seed locations corresponding to the lowest gradient position in 3 x 3 neighbourhood. 
\begin{equation}
G(x,y)=\norm{I(x+1, y)-I(x-1, y)}^{2} + \norm{I(x, y+1)-I(x, y-1)}^{2}
\end{equation}
where I(x,y) is the rgb or lab vector corresponding to the pixel at position (x,y), and ${\norm{.}}$ is the L2 norm. This takes into account both color and intensity information. Each pixel in the image is associated with the nearest cluster center whose
search area overlaps this pixel. After all the pixels are associated with the nearest cluster center, a new center is computed as the average vector of all the  pixels  belonging  to  the  cluster.  We  then  iteratively  repeat  the  process  of
associating pixels with the nearest cluster center and recomputing the cluster
center until convergence. We enforce connectivity in the last step of our algorithm by relabeling disjoint segments
with the labels of the largest neighboring cluster. We enforce
connectivity in the last step of our algorithm by relabeling disjoint segments
with the labels of the largest neighboring cluster. This step is O(N) complex
and takes less than 10${\%}$ of the total time required for segmenting an image.\\
\begin{algorithm}[H]
\SetAlgoLined
\begin{enumerate}
\item Initialize cluster centers ${C_k = [l_k , a_k , b_k , x_k , y_k ]^{T}}$ by sampling pixels at regular grid
steps S.
\item Perturb cluster centers in an ${n × n}$ neighborhood, to the lowest gradient position.
\item \textbf{repeat}
\item \qquad \textbf{for} each cluster center $C_{k}$ \textbf{do}
\item \qquad \qquad Assign the best matching pixels from a ${2S*2S}$ square neighborhood 
\item[] \qquad \qquad around the cluster center according to the distance measure (Eq. 1).
\item \qquad \textbf{end for}
\item \qquad Compute new cluster centers and residual error E {L1 distance between prev\item [] \qquad ious
 centers and recomputed centers}
\item \textbf{until} E ≤ threshold
\item  Enforce connectivity.
\end{enumerate}
 \caption{SLIC Superpixel segmentation}
\end{algorithm}
\subsection{Time Complexity}
By virtue of using our distance measure of Eq. (1), we are able to localize our pixel search on the image plane that is inversely proportional to the number of superpixels K. A pixel falls in the local neighborhood of no more than eight cluster
centers. The convergence error of our algorithm drops sharply
in a few iterations. The time complexity for the classical k-means algorithm is O(N KI) where N is the number of data points (pixels in the image), K is the number of clusters (or seeds), and I is the number of 	iterations required for convergence. But SLIC superpixel segmentation has time complexity of O(N) since we need to compute distances from
any point to no more than eight cluster centers and the number of iterations is
constant. 
\begin{figure}[H]
\centering
\includegraphics[width=80mm]{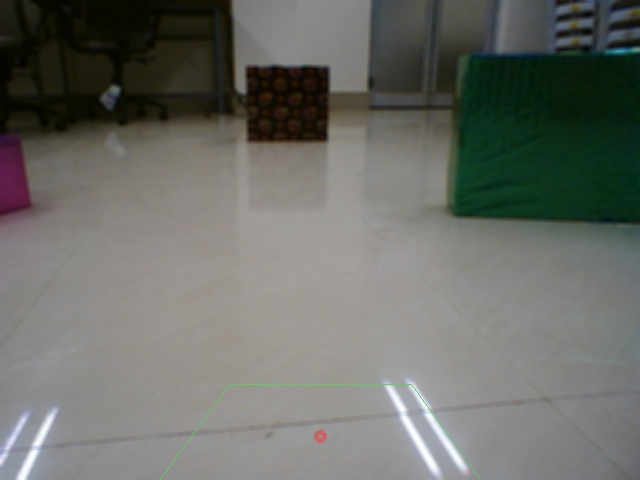}

\caption{Camera Image}
\end{figure}
\begin{figure}[H]
\hfill
\subfigure[K=10]{\includegraphics[width=72mm]{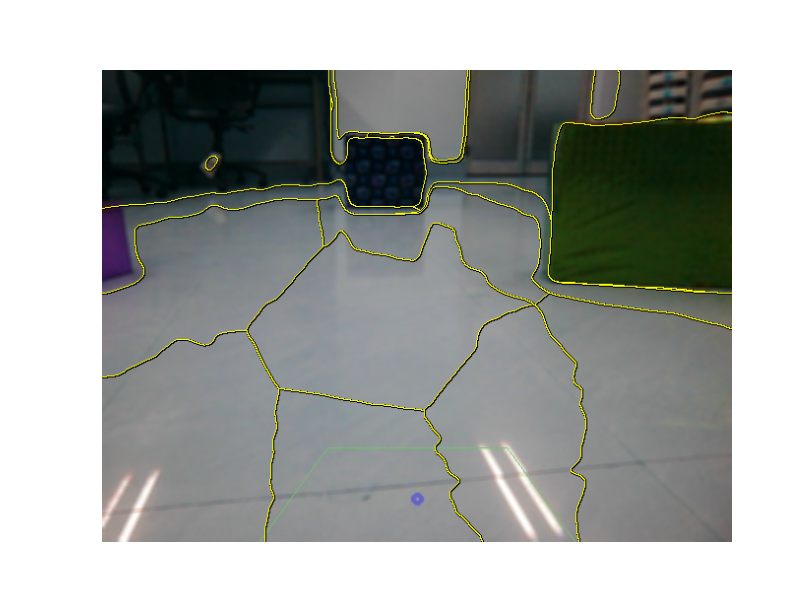}}
\hfill
\subfigure[K=100]{\includegraphics[width=72mm]{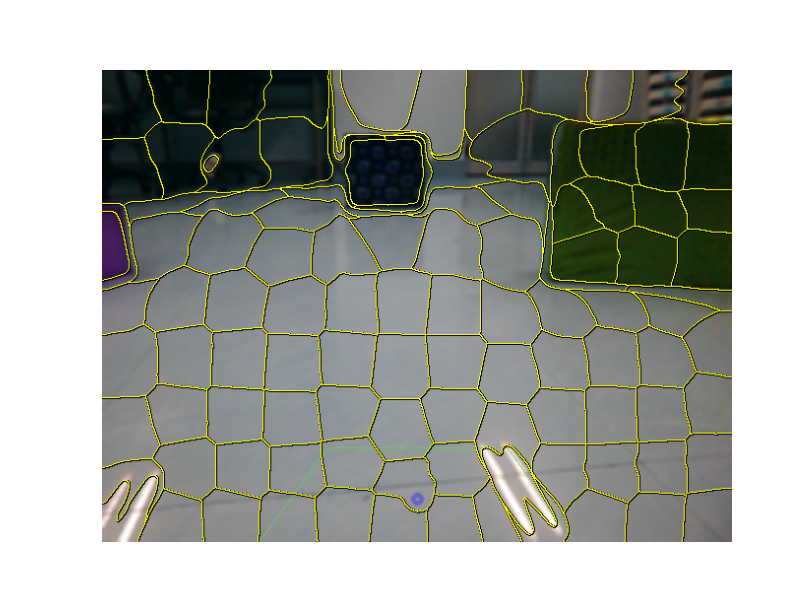}}
\hfill
\subfigure[K=200]{\includegraphics[width=72mm]{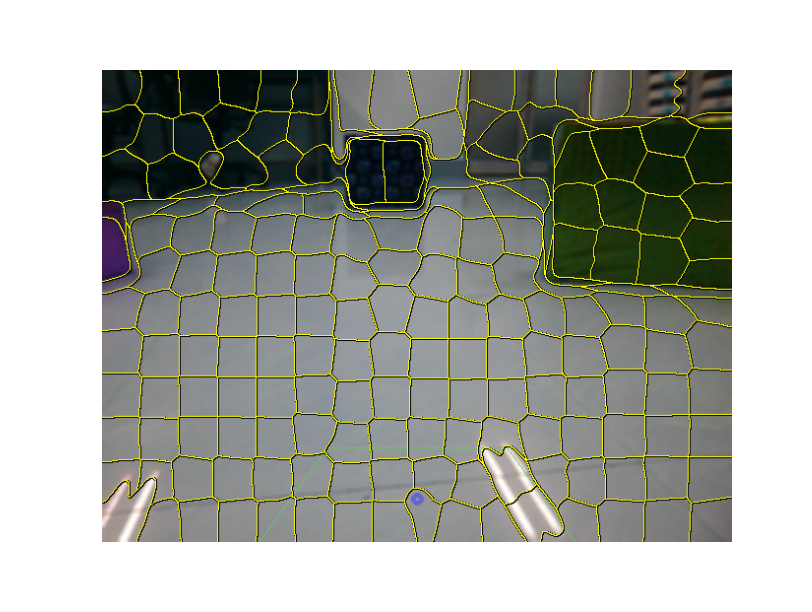}}
\hfill
\subfigure[K=300]{\includegraphics[width=72mm]{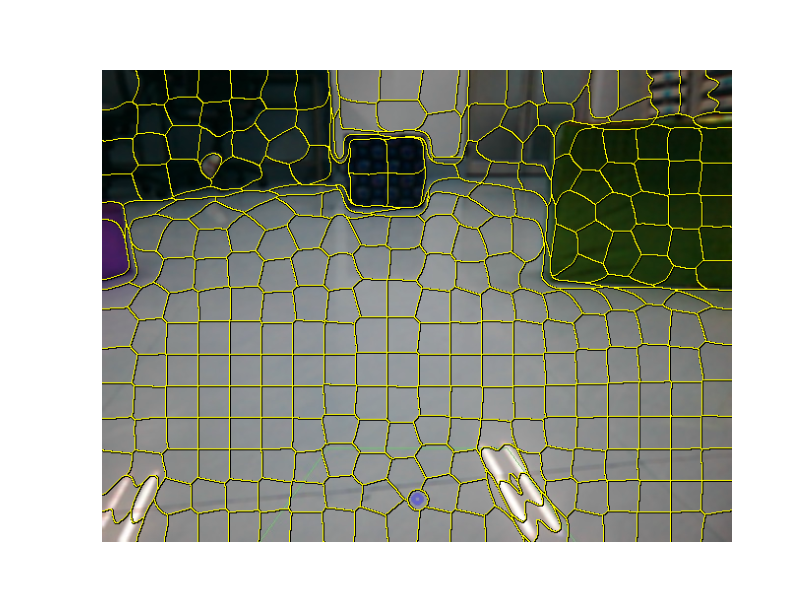}}
\caption{After applying SLIC superpixel segmentation}
\end{figure}

\chapter{Preprocessing of Image}

\section{Color Space}
A color space is the type and number of colors that originate from the combinations of color components of a color model. A color model is an abstract configuration describing how color impression can be created, which consists of color components and rules about how these components interact. 

SLIC superpixel segmentation was applied to RGB image and after converting it to lab color space.
\subsection{RGB color space}
RGB color space is defined by three choromaticities of red, green and blue primaries.
Fig 4.1 shows the superpixel segementation on RGB image.
\begin{figure}[H]
\centering
\includegraphics[width=90mm]{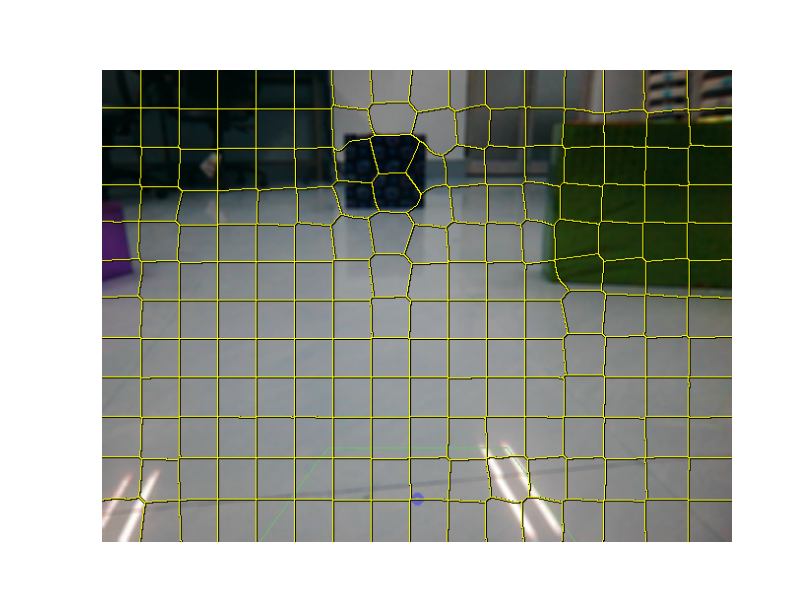}

\caption{Superpixel segmentation for K=200 in rgb color space.}
\end{figure}

\subsection{Lab color space}
A Lab color space is a color-opponent space with dimension L for lightness and a and b for the color-opponent dimensions, based on nonlinearly compressed coordinates. CIE L*a*b* (CIELAB) is a color space specified by the International Commission on Illumination. It describes all the colors visible to the human eye.
The three coordinates of CIELAB represent the lightness of the color (L* = 0 yields black and L* = 100 indicates diffuse white), its position between red/magenta and green (a*, negative values indicate green while positive values indicate magenta) and its position between yellow and blue (b*, negative values indicate blue and positive values indicate yellow). 
Fig 4.2 shows the superpixel segmentation after conversion to lab color space.
\begin{figure}[H]
\centering
\includegraphics[width=90mm]{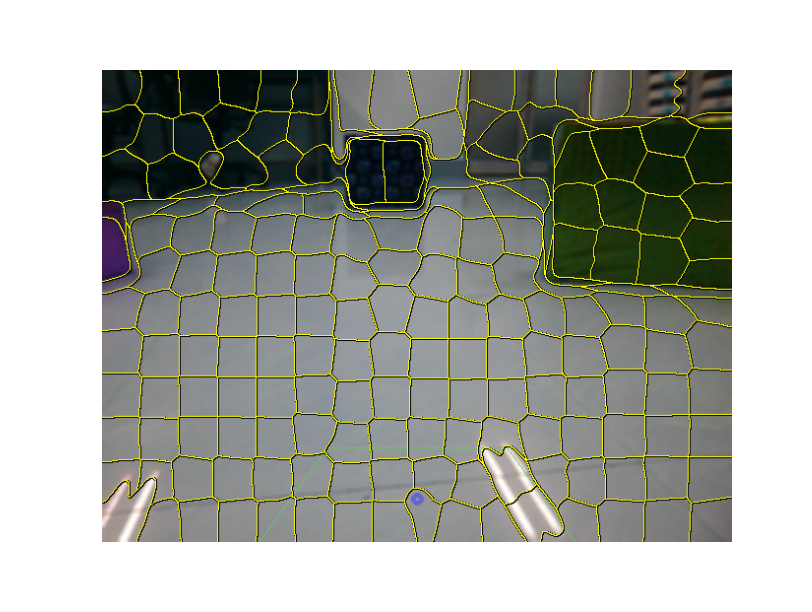}

\caption{Superpixel segmentation for K=200 in lab color space.}
\end{figure}

\section{Gaussian Kernel Smoothening}

The image taken by camera is inherently noisy due to errors associated with image acquisition. Gaussian Kernel smoothening technique is used to remove noise from our image before further processing.
Width of Gaussian smoothing kernel for pre-processing for each dimension of the image. The same sigma is applied to each dimension in case of a scalar value. The
Gaussian kernel
in 1D is defined as:
\begin{equation}
K(t)=\frac{1}{\sqrt{2\pi }}e^{t^{2}/2}
\end{equation}
After scaling the Gaussian Kernel K by the bandwidth sigma ${(\sigma)}$:
\begin{equation}
K_{\sigma}(t)=\frac{1}{\sigma }K(\frac{t}{\sigma })
\end{equation}
This is the density function of the normal distribution with mean 0 and variance ${\sigma^{2}}$.

For our experiment we have taken ${\sigma}$ value as 5.
The n-dimensional isotropic Gaussian kernel is defined as the product of n 1D kernels. Let ${t=(t_1,....,t_n)^{'}\in{\mathbb{R}^{n}} }$. Then the n-dimensional kernel is given by:
\begin{equation}
K_{\sigma}(t)=K_{\sigma}(t_1)K_{\sigma}(t_2)K_{\sigma}(t_3).....K_{\sigma}(t_n)
\end{equation}
\begin{equation}
       =\frac{1}{(2\pi )^{n/2}\sigma ^{n}}\exp (\frac{1}{2\sigma ^{2}}\sum_{i=1}^{n}t_{i}^{2})
\end{equation}

Without smoothening:
\begin{figure}[H]
\centering
\includegraphics[width=90mm]{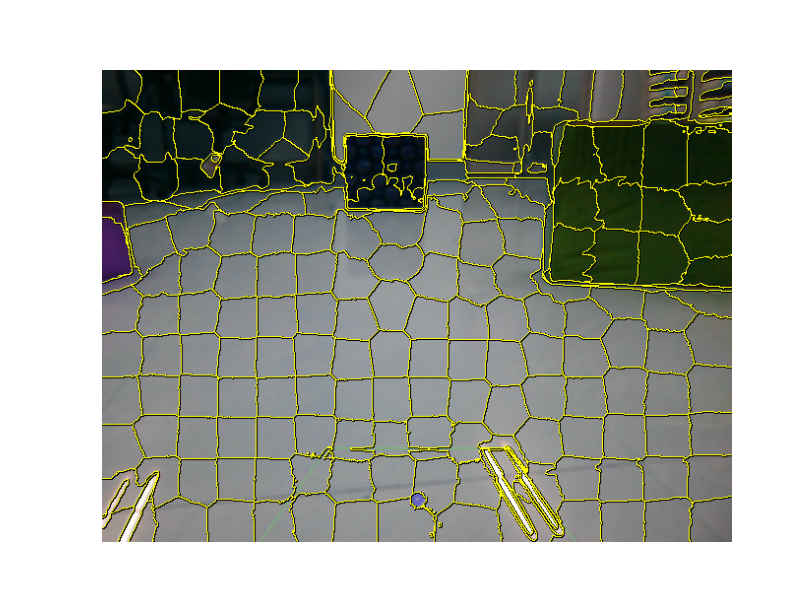}

\caption{K=200}
\end{figure}

After smoothening:

\begin{figure}[H]
\centering
\includegraphics[width=90mm]{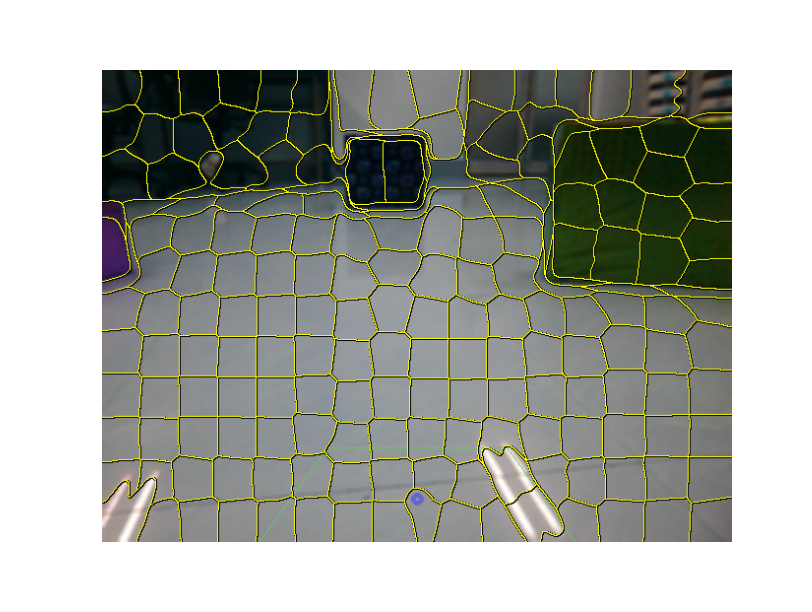}

\caption{K=200, ${\sigma}$=5}
\end{figure}


\chapter{Histogram Sampling}

In our experiment, we have defined safe zone as the region directly in front of the robot that is free of obstacles. We have used a fixed trapezoidal safe zone as shown in Fig \ref{one}.

Histogram gives us a rough idea of what the rest of
the traversable region should look like, because the area directly in front of the robot
is generally a good indication of what the rest of the traversable area could look like.

\section{Accessing individual superpixel of segmented image}

After applying SLIC superpixel segmentation, we get a 2D segments array of same width and height as the original image. Furthermore, each segment is represented by a unique integer, meaning that pixels belonging to a particular segmentation will all have the same value in the segments  array. We construct a mask of the same width and height a the original image and has a default value of 0 (black).
 By stating  segments = segVal  we find all the indexes, or (x, y) coordinates, in the segments  list that have the current segment ID, or segVal . We then pass this list of indexes into the mask and set all these indexes to value of 255 (white).
Applying these masks we will get individual superpixel which we can use for further processing.
\begin{figure}[H]
\centering
\includegraphics[width=100mm]{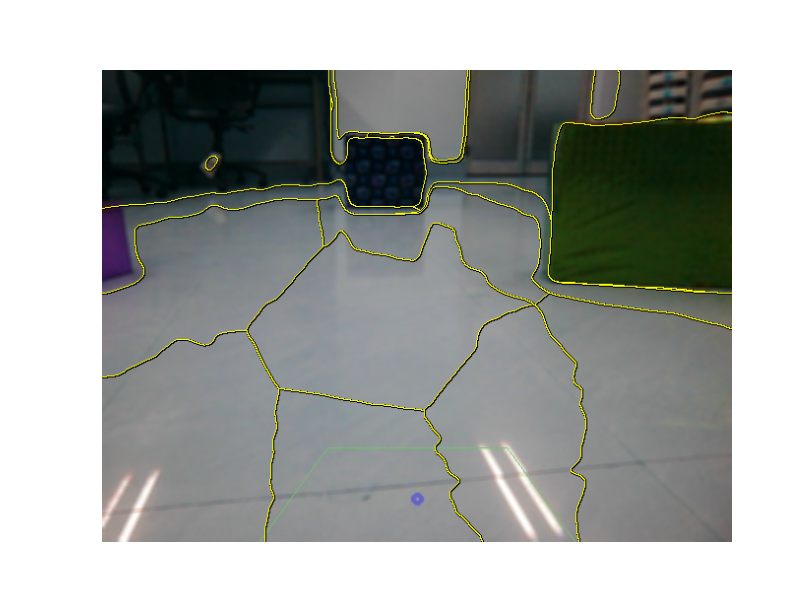}

\caption{SLIC superpixel segmentation for K=10}
\label{one}
\end{figure}

\begin{figure}[H]
\hfill
\subfigure[Mask]{\includegraphics[width=72mm]{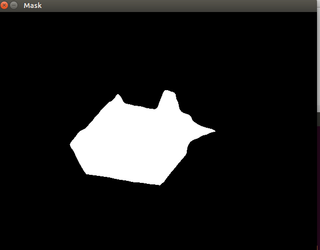}}
\hfill
\subfigure[Accessing the corresponding superpixel after applying mask]{\includegraphics[width=70mm]{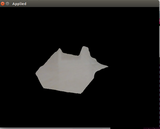}}
\caption{Accessing individual pixel after SLIC segmentation for K=10.}
\end{figure}

\section{Histogram}
The histogram plots the number of pixels in the image (vertical axis) with a particular brightness value (horizontal axis). A histogram represents the distribution of colors in an image \cite{histogram}. It can be visualized as a graph (or plot) that gives a high-level intuition of the intensity (pixel value) distribution. We are going to assume a RGB color space in this example, so these pixel values will be in the range of 0 to 255.

The RGB histogram is shown in figure:
\begin{figure}[H]
\centering
\includegraphics[width=100mm]{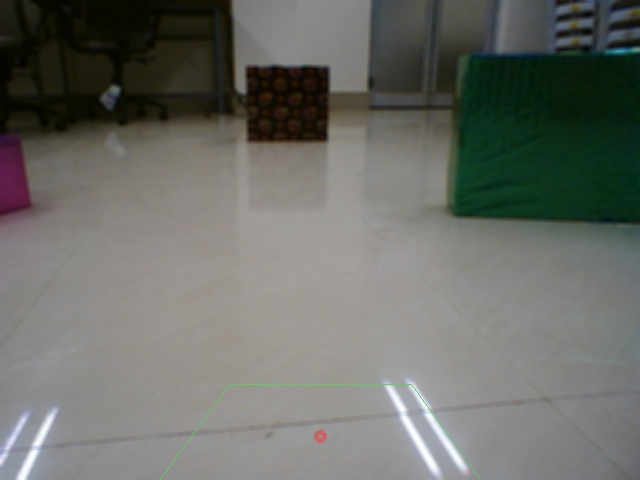}

\caption{Camera Image}
\end{figure}
\begin{figure}[H]
\centering
\includegraphics[width=150mm]{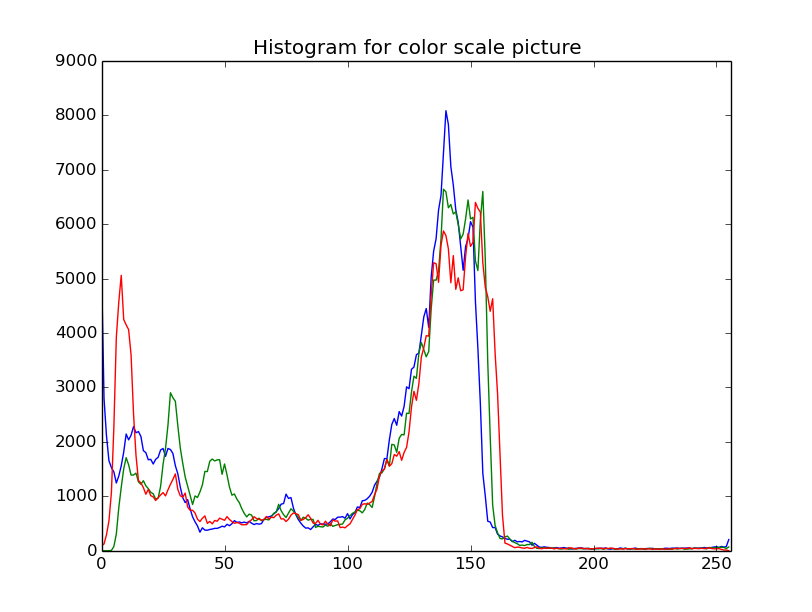}

\caption{RGB histogram of camera image}
\end{figure}

\subsection{Safe zone histogram sampling}
For safe zone histogram sampling we use the following algorithm 2.
\begin{algorithm}[H]
\SetAlgoLined
\textbf{Input}: superpixel image, safe zone coordinates\\
\textbf{Output}: BGR histogram
\begin{enumerate}

\item Split input image into constituent B, G, R planes.
\item Create 3 empty histograms: B, G, R.

\item For each superpixel that has a non-empty intersection with the the safe zone, ob-
tain its B, G and R components and increment the corresponding intensities in the
respective histograms.
\item \textbf{return} BGR histogram

\end{enumerate}
 \caption{SLIC Superpixel segmentation}
\end{algorithm}
\begin{figure}[H]
\centering
\includegraphics[width=100mm]{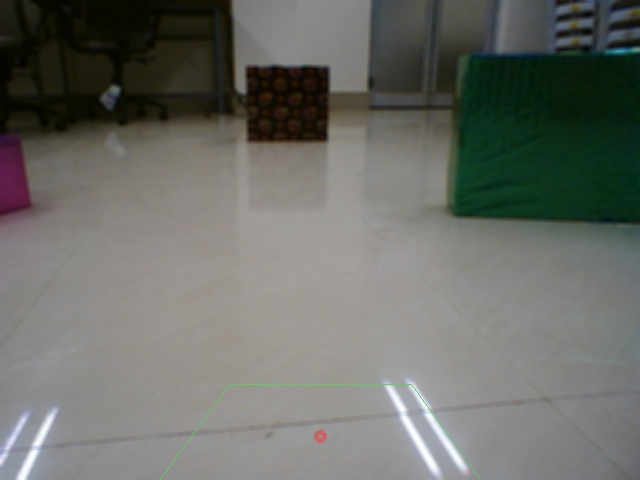}

\caption{The trapezoidal safe zone, shown in green is used as safe zone.}
\end{figure}

\chapter{Floor Segmentation}
Our main aim is to use pixels instead of superpixels to reduce the cost of computation (\cite{bayes} and \cite{segmentation}). In order to segment the floor from other few observations are made. We start
by defining a dependent variable which value will indicate
the probability that a superpixel belongs to the floor. The
training phase uses the value of this dependent variable for
a small set of superpixels that are considered part of the
floor.

\section{Calculation of independent Variables}
In order to segment the floor from the other objects in
an indoor scene we use the following observations:
\begin{itemize}
    \item We have observed that the shapes of the superpixels area near object
boundaries are not regular 
    \item We assume that the superpixels in the safe zone of the images captured by our robot will always correspond to the floor
    \item The color of the superpixels (i.e., the center pixel in the
superpixel area) belonging to the floor should be very
similar.
    \item The shape (bounding box) of the superpixels area that
contains pixels with very similar texture tends to be
regular, like a square
\end{itemize}

Using the above observations, following independent parameters are calculated for each superpixel:
\begin{itemize}
    \item L, a, b channels of superpixels (3 variables)
    \item actual area of superpixels (1 variable)
    \item the width, height and diagonals measures of the
superpixel area (3 variables)
\end{itemize}

Therefore a total of 7 independent variables are used for differentiating the floor pixel from others. Previously we have already calculated the superpixel segmented image using SLIC. All the processing and calculations are done in cielab color space. (L, a, b) channel values of each superpixel centers is calculated by iteration. Actual area will be proportional to total number of pixels in one superpixel. Width and height are calculated by noting the maximum horizontal and vertical perpendicular distance from center, i.e number of pixels in both direction with same label. Similarly, length of diagonal is calculated by traversing 45 degree in both direction from center and summing the number of pixels with same label.
\begin{figure}[H]
\centering
\includegraphics[width=80mm]{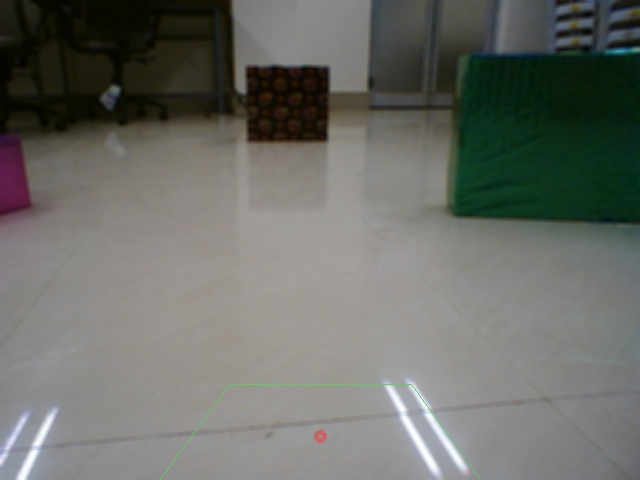}
\caption{Camera Image}
\end{figure}
\begin{figure}[H]
\centering
\includegraphics[width=80mm]{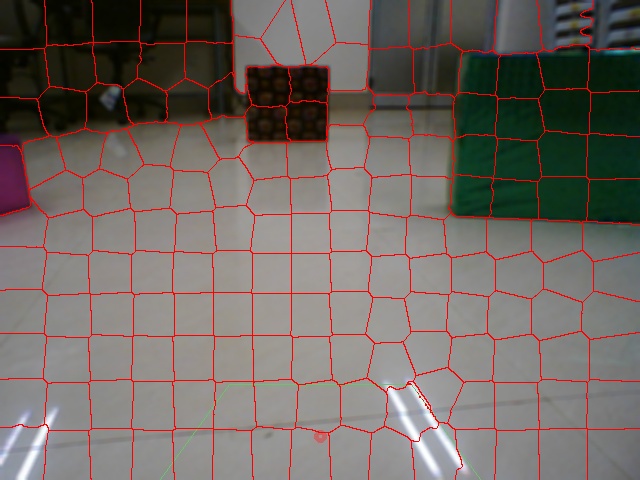}
\caption{After applying SLIC for k=200}
\end{figure}
\begin{figure}[H]
\centering
\includegraphics[width=80mm]{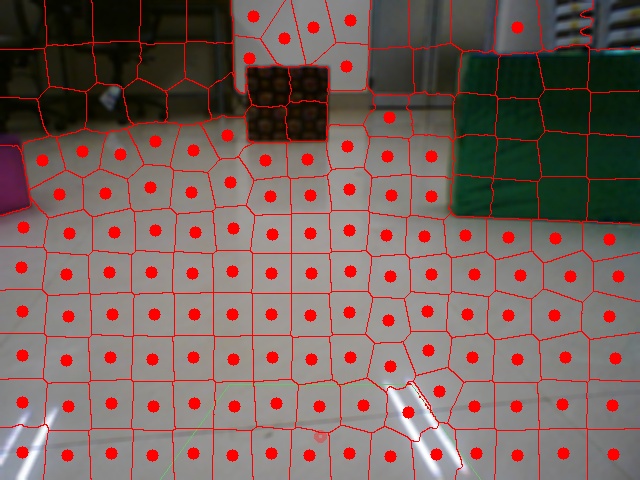}
\caption{After applying floor segmentation}
\end{figure}
\section{Classification of Superpixel}
We use a normalized SSD(Sum of Squared Difference) measure to classify if a superpixel belongs to the floor or not. We first compute the mean values for each independent variable of the superpixels in the training set. Then these mean values are subtracted from each independent variable in a superpixel to classify. For the classification, we define a threshold value according to the maximum normalized SSD measure obtained from
the training set. For (width, height, diagonal, $l, a, b,$ area) as $(a, b, c, d, e, f, g)$ in every superpixel we calculate:\\
${ssd =  (a-a_{mean})^{2}+(b-b_{mean})^{2}+(c-c_{mean})^{2}+(d-d_{mean})^{2}+(e-e_{mean})^{2}+(f-f_{mean})^{2}}$\\
   ${+(g-g_{mean})^{2}}$
\begin{algorithm}[H]
\SetAlgoLined
\begin{enumerate}
\item Convert the image in Lab space.
\item Perform SLIC on image.
\item For every superpixel calculate all 7 parameters.
\item Find the set of supepixel coinciding with safe zone of image.
\item For coinciding suerpixels, find the mean value of all 7 parameters.
\item Subtract the mean value of each parameter from each superpixel's parameter value.
\item Find sum of squared difference calculated above.
\item Using the safe zone superpixels, find the threshold value.
\item Compare the ssd value of each superpixel with threshold value and categorize as them as floor and non-floor.

\end{enumerate}
 \caption{SLIC based floor segmentation (image):}
\end{algorithm}

\subsection{Experimental Results}
Since we require real-time navigation, we focus mainly
on reducing as much as possible the computational time of
our segmentation algorithm while keeping its robustness.
The computational time for 480 x 640 image is approximately 5 seconds. First, we analyze the effect of using the CIE Lab color
space in our classification since this is the color space the
SLIC superpixels use. We note that false changes in intensity
caused by peculiarities in the floor are kept mostly by the
luminance (L) channel. These peculiarities are present due
to illumination conditions and properties of the floor itself.

Superpixels with a red dot in its center indicate that all
pixels in their area belongs to the floor while superpixels
with a no dot are considered as no-floor.  Figure 5 shows a set of different indoor
images. Our approach achieves nearly 90\% detection of free
space on the images in our database. The segmentation is
good even on highly textured floors and when specularities
are present.

\begin{figure}[H]
\hfill
\subfigure[]{\includegraphics[width=70mm]{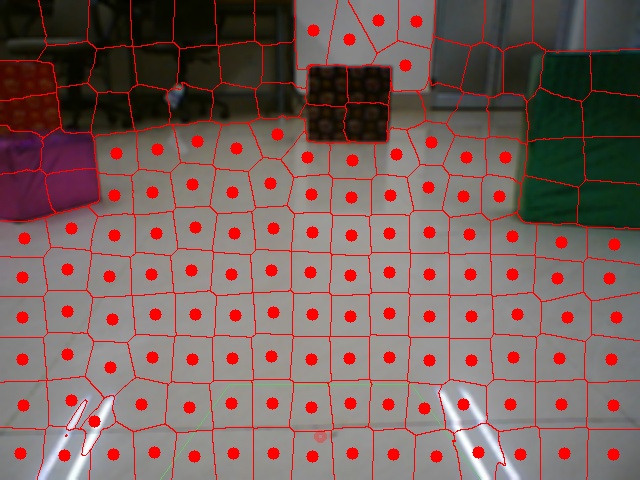}}
\hfill
\subfigure[]{\includegraphics[width=70mm]{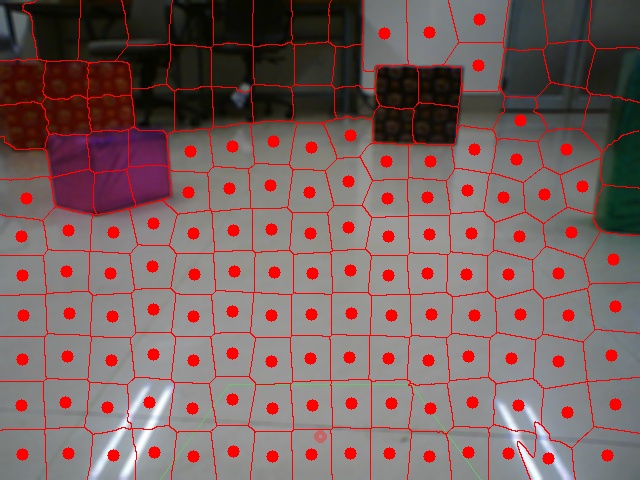}}
\hfill
\subfigure[]{\includegraphics[width=70mm]{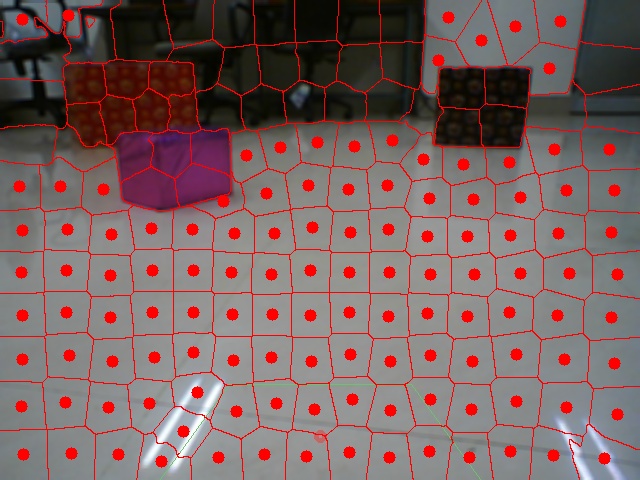}}
\hfill
\subfigure[]{\includegraphics[width=70mm]{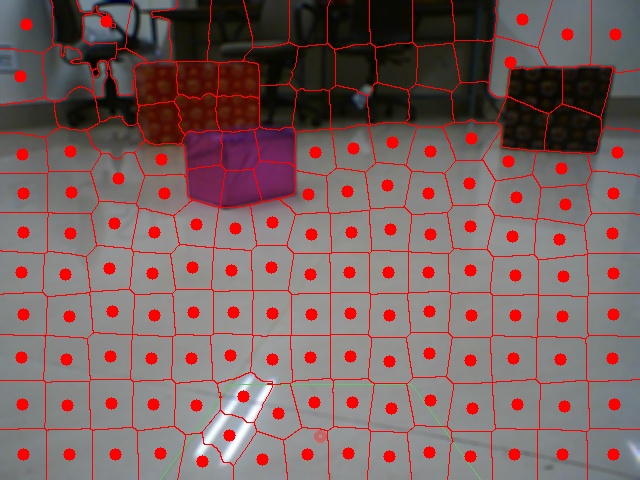}}
\hfill
\subfigure[]{\includegraphics[width=70mm]{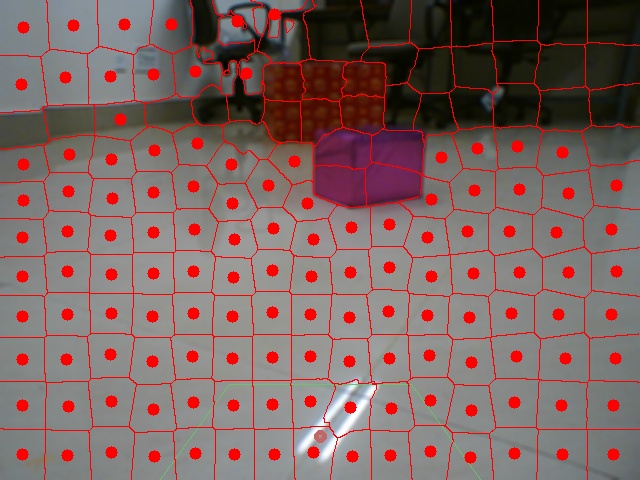}}
\hfill
\subfigure[]{\includegraphics[width=70mm]{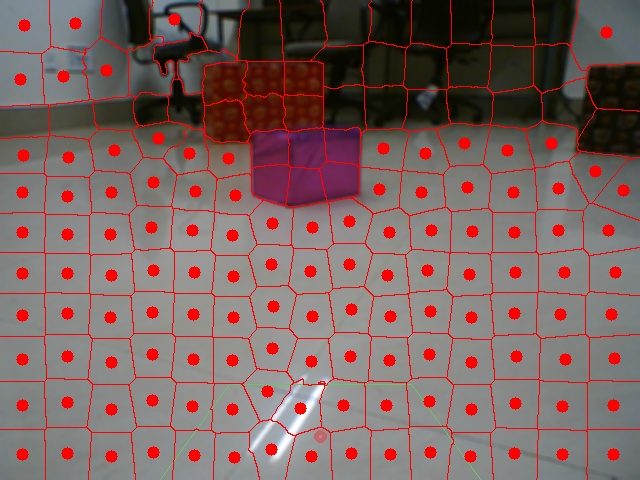}}
\caption{Applying floor segmentation on different images.}
\end{figure}

\section{Problem}
The superpixels of walls of room having the same texture as that of floor are marked as floor. We have ignored the fact that the upper region of the image are less probable to be part of the floor and have not used the superpixel center coordinates while computing SSD. This problem is tackled later by using hough lines to find the walls.

\section{Conclusions}
We have demonstrated that with only one monocular camera and
low resolution images the detection of free space can be
robustly achieved. Our approach uses the SLIC superpixels
to initially segment the input low resolution image and a
normalized SSD similarity measure to classify superpixels
that belongs to the floor (free space). The results shown
that even with specular reflections, shadows coming from far
located objects, and small objects on the floor, our method
efficiently segments the free space.

\section{Other floor segmentation methods}
Flood fill, also called seed fill, is an algorithm that determines the area connected to a given node in a multi-dimensional array. 
The flood-fill algorithm takes three parameters: a start node, a target color, and a replacement color. The algorithm looks for all nodes in the array that are connected to the start node by a path of the target color and changes them to the replacement color. There are many ways in which the flood-fill algorithm can be structured, but they all make use of a queue or stack data structure, explicitly or implicitly.
The main problem with this method is that it is pixel based computation also the varying illumination condition is not handled.
The results are shown in Fig 5.5. Seed is marked as red color circle in safe zone of camera image in Fig 5.1.

\begin{algorithm}[H]
\SetAlgoLined
\begin{enumerate}
\item If target-color is equal to replacement-color, return.
\item If the color of node is not equal to target-color, return.
\item Set the color of node to replacement-color.
\item Perform Flood-fill (one step to the south of node, target-color, replacement-color).
\item Perform Flood-fill (one step to the north of node, target-color, replacement-color).
\item Perform Flood-fill (one step to the west of node, target-color, replacement-color).
\item Perform Flood-fill (one step to the east of node, target-color, replacement-color).
\item Return.

\end{enumerate}
 \caption{Flood-fill (node, target-color, replacement-color):}
\end{algorithm}
\begin{figure}[H]
\hfill
\subfigure[]{\frame{\includegraphics[width=70mm]{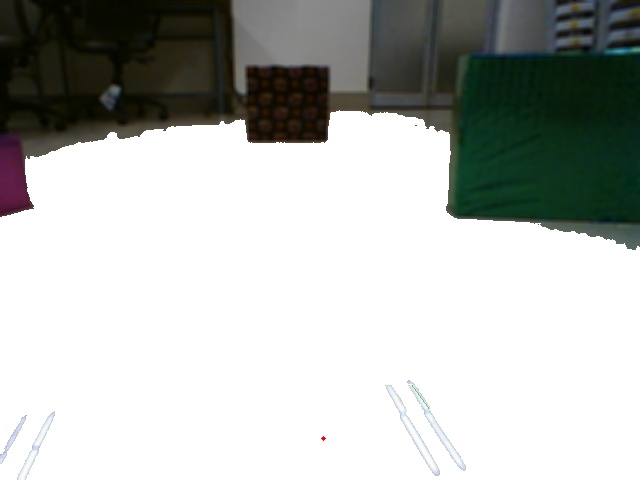}}}
\hfill
\subfigure[]{\frame{\includegraphics[width=70mm]{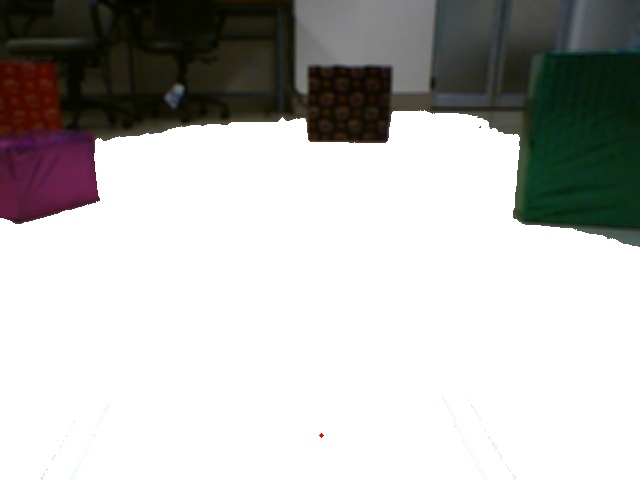}}}
\caption{Applying Flood fill on different images.}
\end{figure}

\section{Floor Junction Masking}
We have seen the problems faced while floor segmentation due to wll superpixels. So in order to further improve the occupancy image, this algorithm detects probable floor junctions
using Canny Edge Detection, contour detection using Suzuki85 algorithm 
and the Probabilistic Hough Line transform to mask the image, and get even better results. The key is using vertical lines as
possible edges of walls/obstacles, since vertical lines remain vertical (invariant) no matter
how the robot is oriented (keeping in mind the stated assumptions) whereas this is not
the case with horizontal lines. For each of the detected vertical lines, the algorithm
estimates where it meets the floor by finding the non-vertical lines that lie within a
threshold radius from the bottom of the vertical line under consideration. Accordingly,
it masks the corresponding region in the occupancy map. The approach is outlined in following Algorithm.

Note that this stage operates directly on
the smoothed camera image, not on the superpixels.
The mask produced by Algorithm is ANDed with the occupancy image produced by
previous algorithms applied in succession to give the final result.
\begin{figure}[H]

\subfigure[]{\frame{\includegraphics[width=60mm]{13_1_cam.jpg}}}
\hfill
\subfigure[]{\frame{\includegraphics[width=60mm]{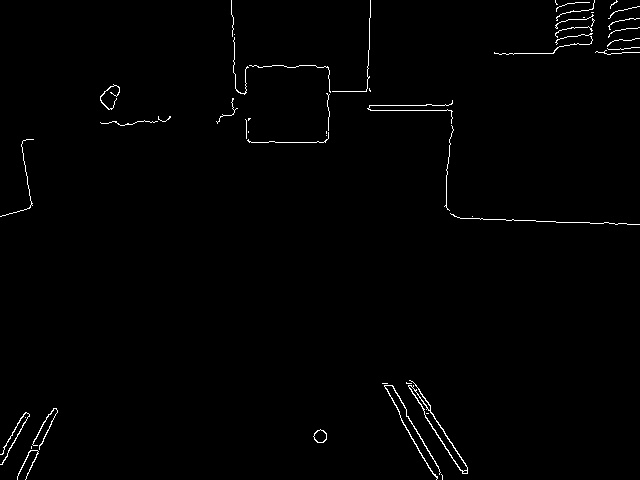}}}
\caption{Applying Canny edge detection}
\end{figure}

\begin{algorithm}[H]
\SetAlgoLined
\textbf{Input:} camera image
\textbf{Output:} binary occupancy mask
\begin{enumerate}
\item Init occupancy mask m ${\leftarrow{}}$ WHITE for all pixels.
\item Init floor junctions list F J ${\leftarrow{}\left \{  \right \}}$.
\item Convert input image from BGR${\rightarrow{}}$  Grayscale.
\item Detect edges E in input image use Canny edge detection.
\item Detect contours C from E using Suzuki85 algorithm.
\item Obtain lines L from C using Probabilistic Hough Transform.

\item \textbf{for all} vertical lines vl in L \textbf{do}
\item  \qquad \textbf{for all} non-vertical lines f l in L \textbf{do}
\item \qquad \qquad \textbf{if}f l lies within circle of threshold radius from bottom of vl \textbf{then}
\item  \qquad \qquad \qquad F J.add((vl, f l))
\item \qquad \qquad  \textbf{end if}
\item \qquad \textbf{end for}
\item \textbf{end for}
\end{enumerate}
 \caption{Floor Junction Masking}
\end{algorithm}

\begin{figure}[H]

\subfigure[]{\frame{\includegraphics[width=60mm]{13_1_cam.jpg}}}
\hfill
\subfigure[]{\frame{\includegraphics[width=60mm]{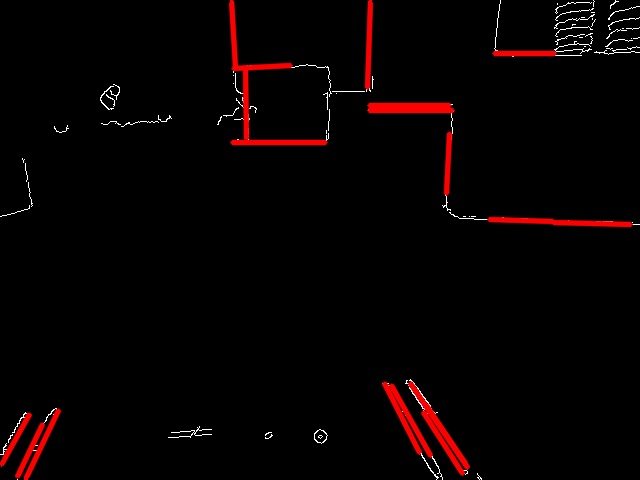}}}
\hfill
\subfigure[]{\frame{\includegraphics[width=60mm]{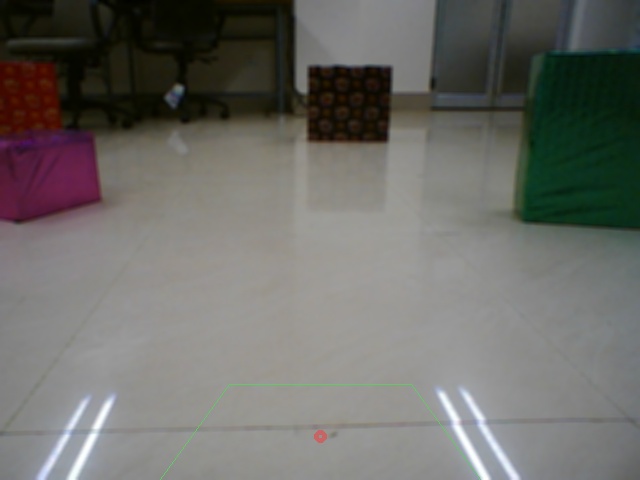}}}
\hfill
\subfigure[]{\frame{\includegraphics[width=60mm]{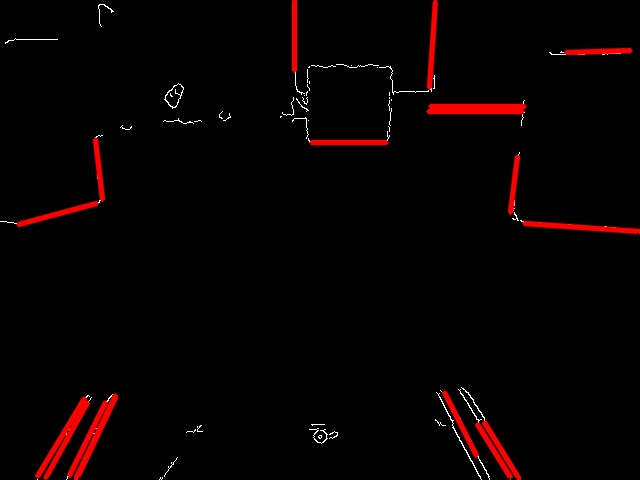}}}
\hfill
\subfigure[]{\frame{\includegraphics[width=60mm]{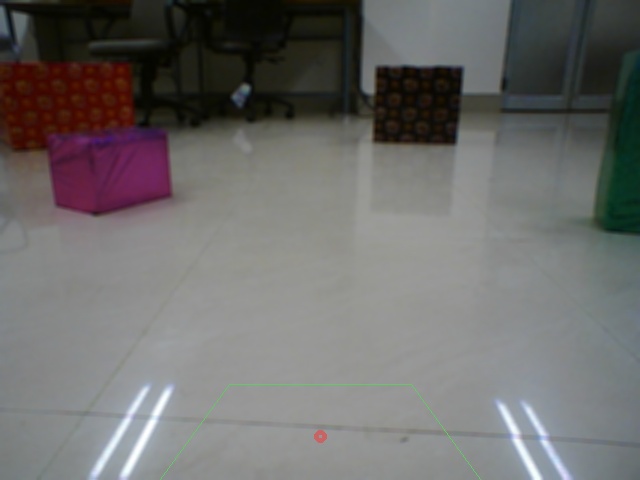}}}
\hfill
\subfigure[]{\frame{\includegraphics[width=60mm]{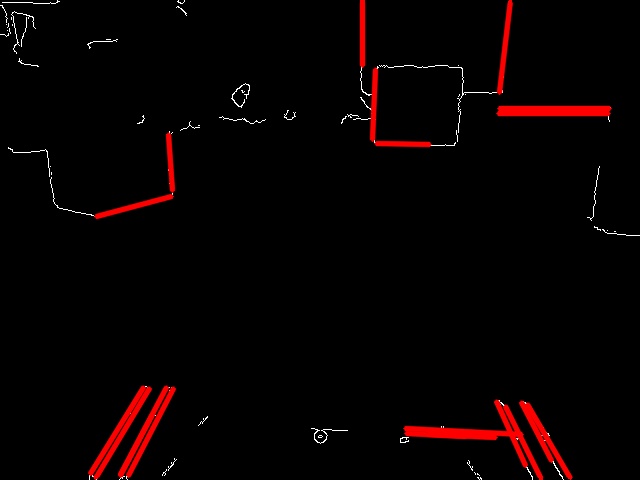}}}
\hfill
\subfigure[]{\frame{\includegraphics[width=60mm]{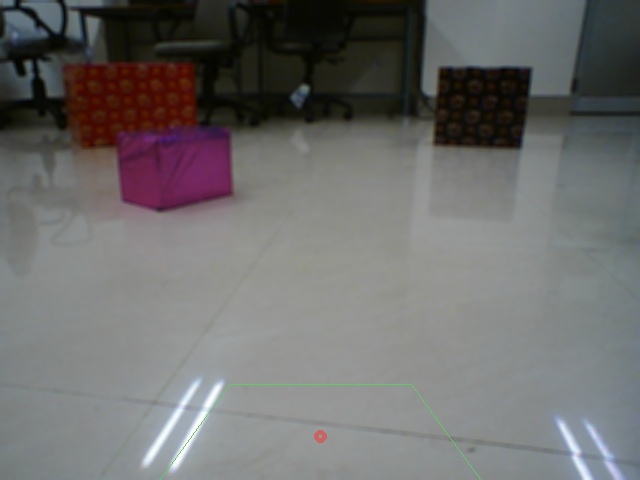}}}
\hfill
\subfigure[]{\frame{\includegraphics[width=60mm]{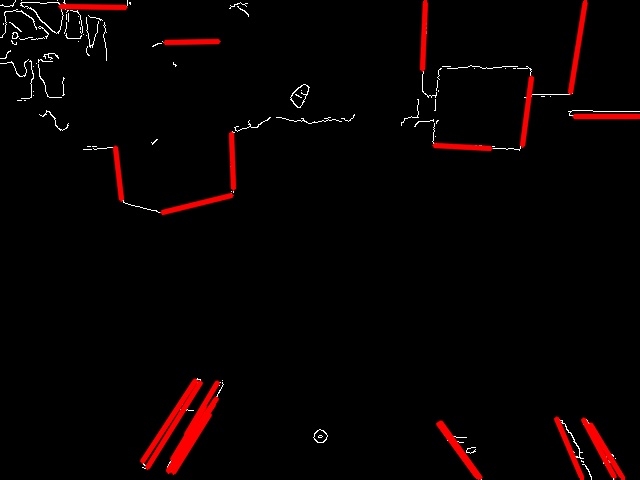}}}
\caption{Applying Hough line transform}
\end{figure}

\section{Region Growing}
Once we get the superpixel segmented region, we apply the region groing to get the floor part only. We define the seed point as the centroid of the safe zone. It is shown in red in Fig. 5.6a.
From the superpixel containing the seed point, we run a breadth first search over the
superpixels. We maintain a set of superpixels called the traversable set, which contains
all the regions that have been detected as belonging to the floor. In the BFS, we consider
the 4-neighbors of a superpixel in the traversable set frontier. We add a neighbor to the
traversable set, if the threshold value discussed before is satisfied. The steps are outlined below.

\begin{algorithm}[H]
\SetAlgoLined
\textbf{Input:} superpixel image, threshold value, seed point, ssd
\textbf{Output:} binary occupancy mask
\begin{enumerate}
\item queue Q ${\leftarrow{}}$ seed superpixel.
\item occupency image occ ${\leftarrow{}}$ BLACK for all superpixels.
\item Convert input image from BGR${\rightarrow{}}$  Grayscale.
\item \textbf{while} Q is not empty \textbf{do}
\item\qquad Superpixel sp  ${\leftarrow{}}$ dequeue(Q)
\item\qquad \textbf{for all} unvisited 4-nbbrs of sp \textbf{do}
\item\qquad\qquad\textbf{if} ssd < threshold
\item\qquad\qquad\textbf{then} 
\item\qquad\qquad\qquad Q  ${\leftarrow{}}$  enqueue(nbr)
\item\qquad\qquad\qquad occ[nbr]  ${\leftarrow{}}$ WHITE
\item\qquad\qquad\textbf{end if} 
\item\qquad \textbf{end for} 
\item \textbf{end while}
\item\textbf{return} occ

\end{enumerate}
 \caption{Region Growing}
\end{algorithm}
\section{Results}
\begin{figure}[H]

\subfigure[]{\frame{\includegraphics[width=60mm]{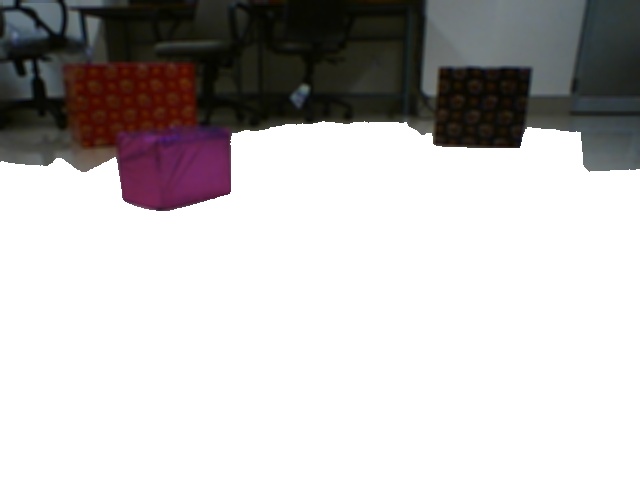}}}
\hfill
\subfigure[]{\frame{\includegraphics[width=60mm]{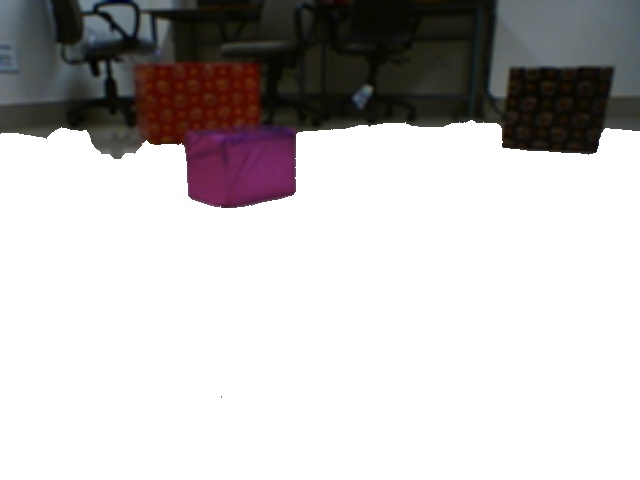}}}
\caption{After applying Region Growing}
\end{figure}
\chapter{Mapping the environment}

\section{Overview}
So  far  in  this  report, we  have  seen  how  to  distinguish  obstacles  from  the  traversal
region just by using monocular vision.  The output of this is a binary occupancy image.
Once a robot can detect obstacles and navigate reliably, it needs to construct a map of
its environment.  Most of the well-established techniques, such as occupancy grids, rely
on range data.  This data is in the form of radial distance and angle measurements to
the obstacles relative to the position of the robot.
However,  for  monocular  vision,  we  need  to  use  a  different  approach  to  calculate  the
range information, since it is not readily available.  The goal of the personal robot is to
dynamically create a map of the unstructured environment that it has been placed.  Having identified the obstacles in
the previous section, placing them on the map requires a transformation from the image
coordinate space to the real world coordinate space.
In  this  section, we describe a method to convert from image plane to ground plane. This is in contrast to many monocular approaches that create a sparse
3-D map of features in the environment, such as \cite{exploration} and \cite{3dvision}.  This representation is fast,
intuitive and more importantly, convenient to work with for subsequent stages, like path-
planning. Finally, we present a look up table which can be used for easy transformation.

\section{Perspective Mapping}
\subsection{ Image Formation and Perspective Mapping}
Monocular images are two dimensional and the process of capturing an image loses the
depth information and introduces a perspective mapping, since all the points in the 3-D
space along a ray of light traced from the camera lens will map to the same pixel in the
image.
Since  obstacles  are  present  in  the  environment,  occlusion  prevents  the  camera  from
seeing past the nearest obstacle along the ray.
Inverse perspective mapping (IPM)
is a
method for turning an image back into a 3-D map, i.e.  it tries to reverse the effect of
the perspective mapping.  It involves an analysis based on homogeneous coordinates and
perspective transformation matrix.
\subsection{Homogeneous Coordinates}
Homogeneous coordinates are simply a way of representing N
-dimensional coordinates with N
+ 1 numbers,  i.e.  the point (x;y) in the Cartesian system becomes (X;Y;w)
in  homogeneous  coordinate  system,  where
w
is  an  additional  variable.   Homogeneous
coordinates allow affine transformations (translation, scaling, shearing, rotation etc.)  to
be easily represented using matrix operations.
\subsection{Homography}
One of the uses of homogenous coordinates is in homography.  A homography is a trans-
formation (matrix) from one plane to another.  Two images are related by a homography
if and only if both images are viewing the same plane from a different angle.  IPM is
thus a homography, as shown in Fig.  6.1 .

\begin{figure}[H]
\centering
\includegraphics[width=90mm]{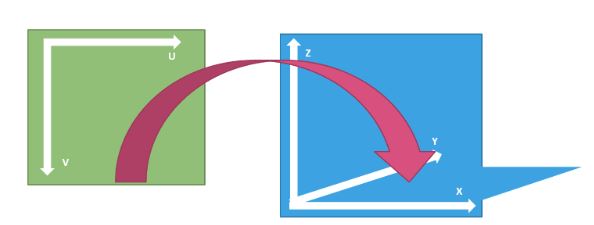}

\caption{Coordinate convention for camera image space (left) and ground space (right)}
\end{figure}

For analysis, we make the assumption that our camera is an idealized pinhole camera,
which suits for most applications.  In that case, the camera matrix, which describes
the mapping from 3D points in the world to 2D points in an image, depends only on the
focal length of the camera.  So, if we knew the focal length, then we could easily  and
this transformation matrix.  However, in general for a digital camera, the focal length is
unknown.  Thus we use another method to estimate the distance using IPM, detailed in
the next section.
\section{Algorithm}
The homography is calculated using 4 point correspondence, giving us all the information to transform between image plane and ground plane coordinates. Limitation of this approach is that it assumes a uniformly parametrized image plane (pinhole camera), so lens distortion will give us errors as seen in my example. If we are able to remove lens distortion effects, we'll go very well with this approach. In addition we will get some error of giving slightly wrong pixel coordinates as our correspondences, we can get more stable values if you provide more correspondences.
Using this input image shown in Fig 6.2.

\begin{figure}[H]
\centering
\includegraphics[width=90mm]{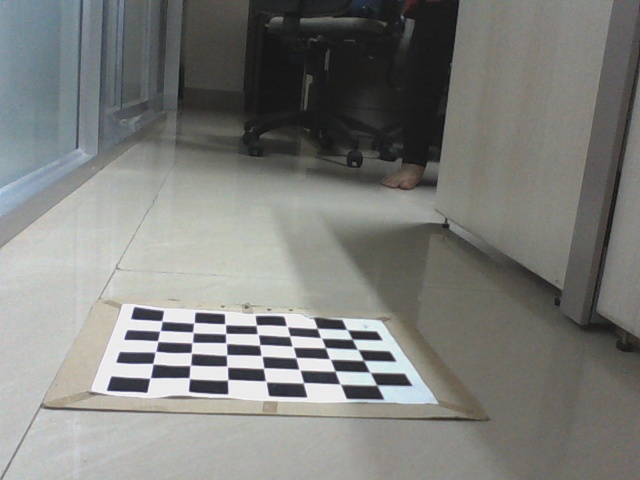}

\caption{Input Image for mapping}
\end{figure}

The four corners of one chess field are read from mouse click, which will correspond to the fact that we know 4 points in our image. Points are marked red in the following image.

\begin{figure}[H]
\centering
\includegraphics[width=90mm]{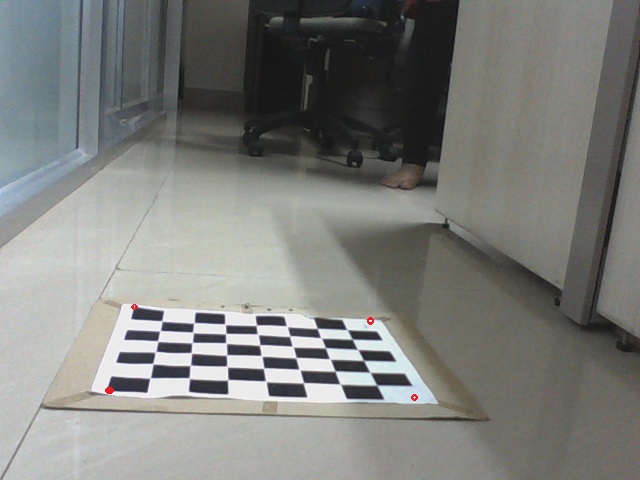}

\caption{Corners of chess field}
\end{figure}

We define 4 points representing our “top-down” view of the image. The first entry in the list is (0, 0)  indicating the top-left corner. The second entry is (maxWidth - 1, 0)  which corresponds to the top-right corner. Then we have (maxWidth - 1, maxHeight - 1)  which is the bottom-right corner. Finally, we have (0, maxHeight - 1)  which is the bottom-left corner.

The takeaway here is that these points are defined in a consistent ordering representation — and will allow us to obtain the top-down view of the image. To actually obtain the top-down, “birds eye view” of the image we’ll utilize the perspective transformation discussed before. For transformation we require the 4 ROI points in the original image and a list of transformed points and we can calculate the transformation matrix M using that. Applying the transformation matrix we get our warped image which is our top-down view. In this image the corner of chess fields are marked.

\begin{figure}[H]
\centering
\includegraphics[width=90mm]{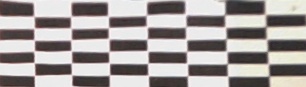}

\caption{Warped Image}
\end{figure}
\begin{figure}[H]
\centering
\includegraphics[width=90mm]{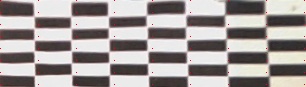}

\caption{Input Image for mapping}
\end{figure}

We can see there is some big amount of error in the data. as said before it's because of slightly wrong pixel coordinates given as correspondences (and within a small area!) and because of lens distortion preventing the ground plane to appear as a real plane on the image. The lookup table is calculated in the same manner for every corners.

Error is calculated using sum of squared difference of the two corresponding points.

\section{Test Platform}

Our test robot is shown in Fig.  6.6.  We have used a Sony Playstation Eye USB webcam
as our sensor, an Arduino Uno as our microcontroller, and a standard laptop processor.
The robot is equipped with a di erential drive mechanism and is powered by a 12 V
battery pack.
The  OpenCV  library,  developed  by  Intel,  has  been  used  to  prototype  and  test  our
algorithms.
\begin{figure}[H]
\centering
\includegraphics[width=90mm]{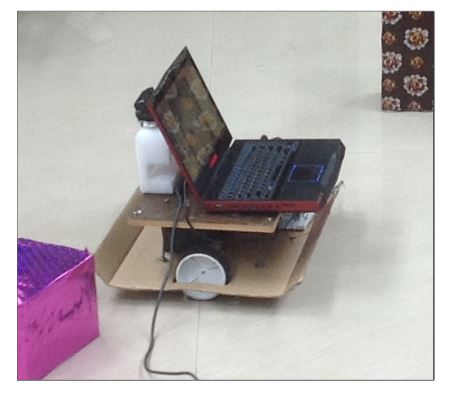}

\caption{Test robot}
\end{figure}

\chapter{Future Work}

Future work which can be carried on are as follows:
\begin{itemize}

\item Probabilistic occupancy grid generation

SLAM algorithm can be combined with an exploration procedure. Exploration strategy should
be able to cover the unknown terrain as fast as possible avoiding repetition as much as possible but this is suboptimal in the context of slam because the robot typically needs to re-visit places to localize itself. There are two standard methods \cite{monoslam} and \cite{exploration2} which can be used for generating probability occupancy grid.
Using the frontier mapping, we need to estimate the best possible direction.

\end{itemize}

\chapter{Conclusion}In this project, we developed methods to develop some of the building blocks for autonomous robot navigation using monocular vision.
We are working on a multi-stage obstacle detection technique based on monocular vision which has its own advantages and disadvantages. We are using computationally fast algorithm for each step hence reducing the complexity. But on the other hand it has a constraint that it cannot be used in low light situation due to sensor limitations.

 Obstacle detection and
avoidance in real time is a complex and computationally expensive.We have used a novel O(N) complexity superpixel segmentation algorithm that is simple to
implement and outputs better quality superpixels for a very low computational
and memory cost. It needs only the number of desired superpixels as the input
parameter. It scales up linearly in computational cost and memory usage.
Using the superpixel segmentation method a complete map of the traversal region is created which will be used for exploration.
 The problem now lies in developing the best strategy for exploration. After detecting the obstacles, we will be able to generate potential destinations and along with the occupancy map, we will be able to explore the entire region.
 
 In conclusion, this project has shown that the simple webcam which is easily available to everyone can be used for indoor robot navigation, reducing the complexity and computation cost. Thus, this approach is of industrial importance thereby simplifying the daily life needs.




\addtocontents{toc}{\vspace{2em}} 

\addtocontents{toc}{\vspace{2em}}  
\backmatter

\label{Bibliography}
\lhead{\emph{Bibliography}}  
\bibliographystyle{ieeetr}  
\bibliography{Bibliography}  
\end{document}